\pdfoutput=1

\documentclass[11pt]{article}

\usepackage[preprint]{acl}

\usepackage{times}
\usepackage{latexsym}
\usepackage{caption}
\usepackage{subcaption}
\usepackage{graphicx}
\usepackage{balance}

\usepackage[T1]{fontenc}

\usepackage[utf8]{inputenc}

\usepackage{microtype}

\usepackage{inconsolata}

\usepackage{xcolor}

\usepackage{amsmath}
\usepackage{amsfonts}
\usepackage[capitalize,noabbrev]{cleveref}

\title{Mamba Knockout for Unraveling Factual Information Flow}

\author{Nir Endy\thanks{\;\;Equal contribution.} \\
  Tel Aviv University \\
  \texttt{nirendy@gmail.com} \\\And
  Idan Daniel Grosbard\footnotemark[1] \\
  Tel Aviv University \\
  \texttt{idangrosbard@mail.tau.ac.il} \\\And
  Yuval Ran-Milo\footnotemark[1] \\
  Tel Aviv University \\
  \texttt{yuv.milo@gmail.com} \AND
  Yonatan Slutzky\footnotemark[1] \\
  Tel Aviv University \\
  \texttt{slutzky1@mail.tau.ac.il} \\\And
  Itay Tshuva \\
  Tel Aviv University \\
  \texttt{itayt122@gmail.com} \\\And
  Raja Giryes \\
  Tel Aviv University \\
  \texttt{raja@tauex.tau.ac.il}\\}

\begin{document}
\maketitle
\begin{abstract}
This paper investigates the flow of factual information in Mamba-based language models. We rely on theoretical and empirical connections to Transformer-based architectures and their attention mechanisms. Exploiting this relationship, we adapt attentional interpretability techniques originally developed for Transformers—specifically, the Attention Knockout methodology—to both Mamba-1 and Mamba-2. Using them, we trace how information is transmitted and localized across tokens and layers, revealing patterns of subject-token information emergence and layer-wise dynamics. Notably, some phenomena vary between Mamba models and Transformer-based models, while others appear universally across all models inspected—hinting that these may be inherent to LLMs in general. By further leveraging Mamba’s structured factorization, we disentangle how distinct “features” either enable token-to-token information exchange or enrich individual tokens, thus offering a unified lens to understand Mamba's internal operations. Our code can be found at \url{https://github.com/nirendy/mamba-knockout}.
\end{abstract}

\section{Introduction}
Understanding how factual information moves through different parts of a language model is an important step toward explaining its outputs. While Transformer-based models and their attention mechanisms have been studied in depth \citep{geva2023dissecting}, less is known about how structured state-space models (SSMs) process and transfer facts across tokens and layers. Recent work has introduced Mamba-based SSM architectures that rival Transformer performance in various settings \citep{gu2024mambalineartimesequencemodeling, waleffe2024empiricalstudymambabasedlanguage,dao2024transformers}. Yet, the internal pathways of factual information within these models remain less explored.

Motivated by theoretical connections showing that selective SSMs can be understood through an attention-like perspective \citep{ali2024hiddenattentionmambamodels,dao2024transformers,zimerman2025explaining,ben-kish2025decimamba}, we draw on methods originally developed for Transformers. Notably, a certain class of SSMs has been proven equivalent to a subclass of linear attention Transformers \citep{pmlr-v119-katharopoulos20a, dao2024transformers}, suggesting that interpretability tools designed for attention can also be applied to SSM-based architectures. By employing the “Attention Knockout” technique \citep{geva2023dissecting}, previously used to map how factual information flows in Transformers, we can similarly isolate and analyze the flow of information in Mamba-1~\citep{gu2024mambalineartimesequencemodeling} and Mamba-2~\citep{dao2024transformers}.

Our results show that Mamba models, despite their differences from Transformers, display certain patterns of information emergence and routing that align with observed attention-based behaviors. Key facts surface within certain tokens and gain prominence at specific layers, mirroring the step-by-step integration of information commonly noted in Transformers \citep{geva2023dissecting}. In contrast, our results also show that other patterns are differentiated between Mamba models and Transformer-based models, with some phenomena even varying within each family.

By leveraging Mamba’s factorized structure, we separate the roles of different “features” in (i) transmitting information between tokens; and (ii) enriching individual tokens independently. This approach not only clarifies how Mamba’s internal operations resemble those of well-studied attention models, but also provides a clearer framework for future analyses of SSM-based language models.

Ultimately, our work extends interpretability beyond the Transformer paradigm, offering a more unified perspective on how language models—whether attention-dominated or grounded in state-space representations—organize factual information. By illuminating these internal processes, we take a step toward a more generalizable understanding of how language models, and specifically Mamba, form and express factual knowledge.

The contributions of this paper are twofold: (i) We extend the Attention Knockout interpretability framework—previously developed exclusively for Transformers—to structured state-space models (SSMs), enabling us to uncover previously unknown parallels and distinctions between Mamba-based and Transformer-based architectures' factual-information dynamics. (ii) We propose a novel `feature knockout' mechanism that takes advantage of the unique structure of SSMs, allowing nuanced interventions and additional insights into how different feature types contribute to model behavior.

\section{Related Work}\label{sec:related}
Selective State-Space Modeling (SSM) was recently introduced by \citet{gu2024mambalineartimesequencemodeling}, rivaling performance of Transformer-based models in various settings \citep{waleffe2024empiricalstudymambabasedlanguage}. An analytical examination of the core operation in selective SSMs by \citet{ali2024hiddenattentionmambamodels,zimerman2025explaining} provided a complementary view of selective SSMs as attention-driven models, allowing the application of attention-based explainability analysis to SSM-based models.
The Structured State-Space Duality (SSD) framework proposed by \citet{dao2024transformers} bridges SSMs and attention layers even further, proving that a specific class of SSMs is equivalent to a subclass of linear attention Transformers \citep{pmlr-v119-katharopoulos20a}. In this work, we employ these approaches to enable the application of attention-based tools and analyses to SSM-based architectures.

Complementary, several studies have analyzed how factual information is stored and extracted in Transformer-based models. \citet{geva2021transformerfeedforwardlayerskeyvalue} show MLP layers function as key-value memories, which extract semantic associations regarding the input in the ultimate layers. \citet{nichani2024understandingfactualrecalltransformers} prove that shallow Transformers have optimal factual storage in their value matrices and MLPs. To assess in greater accuracy where specific memories are stored, \citet{meng2022locating} suggest Causal Tracing - a method for identifying critical MLPs by adding noise to intermediate representations and restoring clean run states at deeper layers. They also show that in order to edit stored facts, a rank-one matrix editing in the MLP of a single, early, layer is sufficient. The work of \citet{meng2023masseditingmemorytransformer} expands the ROME technique for mass editing memory by expanding the editing to multiple consecutive early layers. \citet{hase2023doeslocalizationinformediting} show that early-intermediate layers are in general good candidates for injecting new facts to LMs, and empirically validate a particular type of Causal Tracing that is insightful for fact editing and fact storage. Our work draws its inspiration from \citet{geva2023dissecting} which analyzed how the information from multiple tokens is aggregated to correctly query attributes using \textit{Attention Knockout}. Their analysis reveals that this relies on an initial enrichment process where the subject token extracts relevant attributes in the early layers' MLPs. 

Our work most closely aligns with that of \citet{sharma2024locatingeditingfactualassociations}, whose thorough investigation of factual associations in Mamba-1 featured an analysis using a variant of attention knockout. However, our approach differs from them in several key aspects. \citet{sharma2024locatingeditingfactualassociations} focus on blocking the propagation of information from a single token to all future tokens, whereas we adopt the technique proposed in \citet{ali2024hiddenattentionmambamodels} to remove attention specifically from one token to a single token—more in line with previous work such as \citet{geva2023dissecting}. Though \citet{sharma2024locatingeditingfactualassociations} suggest that such fine-grained blocking may be challenging due to convolution and softmax layers, we demonstrate that a straightforward and natural implementation successfully replicates the phenomena observed in Transformer-based attention knockouts. We also investigate Mamba-2 (where attention knockout has a more direct interpretation) and find consistent patterns in both Mamba-1 and Mamba-2, suggesting that these dynamics may be fundamental across SSM-based and Transformer-based architectures. Finally, beyond leveraging attention parallels, our work introduces a novel “feature knockout” mechanism that exploits the unique structure of SSMs, enabling more nuanced interventions and additional insights into how different feature types contribute to model behavior. 
We inspect additional related works in \cref{app:related_extended}.

\section{Methodology}\label{sec:method}

We now turn to introduce the tools that we use in our work to analyze Mamba. We start by describing the attention knockout mechanism, which was proposed for Transformers, and the SSMs formulation. Then we explain our methodology for employing knockout for SSMs and specifically for Mamba-1 and Mamba-2.

\subsection{Attention Knockout}\label{sec:method:knockout}
We adopt the \textit{Attention Knockout} methodology introduced by \citet{geva2023dissecting} and apply it to identify critical points in the flow of information essential for factual predictions. For successful next-token prediction, a model must process the input tokens so that the next-token can be inferred from the last position. \citet{geva2023dissecting} investigate this process internally by “knocking out” parts of the computation and measuring the effect on the prediction. To this end, they propose a fine-grained intervention on the attention layers, which, they show, serves as a "crucial" module for communicating information between positions. By disrupting critical information transfer through these layers, they demonstrate that factual predictions are constructed in stages, with essential information reaching the prediction position at specific layers during inference. Intuitively, critical attention connections are those whose disruption results in a marked decline in prediction quality. To test whether essential information flows between two hidden representations at a specific layer, the authors zero out all attention connections between them. Formally, given an input sequence of \(N\) tokens, a model consisting of \(L\) consecutive layers, and two positions \( r, c \in [1,N] \) with \( r \leq c \), they prevent the \( c \)th token at layer \(l\) from attending to  \( r \)th token at layer \(l\) by zeroing the attention weights for that layer at index \( c, r\).

\subsection{Selective State Space Models}
\label{seq:selective}
A selective state space model (SSM) of dimension \(n\) is defined as a time-dependent recurrent relation driven by an input signal \(u(t) \in \mathbb{R}\). At each time step \(t\), the model is parameterized by matrices \(A(t) \in \mathbb{R}^{n \times n}\), \(B(t) \in \mathbb{R}^{n\times 1}\), and \(C(t) \in \mathbb{R}^{1 \times n}\) that can depend on the current input \(u(t)\). Formally, the system evolves according to:
\begin{equation*}\label{eq:discrete_ssm}
\begin{aligned}
x(t+1) &= A(t)\,x(t) + B(t)\,u(t), \\
y(t) &= C(t)\,x(t).
\end{aligned}
\end{equation*}
These parameterizations differ between Mamba-1 and Mamba-2. Further details can be found in \citet{gu2024mambalineartimesequencemodeling, dao2024transformers}. However, a unifying property of both models is that the transfer matrix $A(t)$ is parameterized as
\begin{align*}
    A(t) = \bar{A}^{\Delta(t)},\ \bar{A}=\exp(\mathrm{diag}(\alpha)),    
\end{align*}
for some learned parameters \(\Delta(t)\in\mathbb{R}\) and \(\alpha \in [0,1]^n\), a key detail used in \cref{sec:method:feature_knockout}. 

\subsection{Hidden Attention of Mamba-1}\label{sec:method:mamba1}
For Mamba-1, we follow the \textit{hidden-attention} perspective introduced by \citet{ali2024hiddenattentionmambamodels} to implement attention knockout in Mamba. Specifically, the authors utilized the kernel representation of Selective-SSMs and observed that the relationship between any two tokens \(i\) and \(j\) (where \(i \leq j\)) is represented as an entry in the kernel matrix 
\[
\mathbf{M}_{i,j} = Q_i \cdot H_{i,j} \cdot K_j.
\]
Here, \(Q_i = C(i)\), \(H_{i,j} = \prod_{t=i}^j A(t)\), and \(K_j = B(j)\) (see \cref{seq:selective}). To apply knockout between two tokens at a specific layer, we simply assign \(\mathbf{M}_{i,j}=0\).

\subsection{Implicit Linear Attention of Mamba-2}\label{sec:method:mamba2}
In Mamba-2, following \citet{dao2024transformers}, the SSM layer can be interpreted as a masked linear attention mechanism, allowing it to be expressed as a matrix multiplication. Specifically, the SSM layer processes an input tensor \( \mathbf{X} \in \mathbb{R}^{L \times H} \), where \( L \) represents the sequence length and \( H \) the dimensionality of each input token. The layer transforms \( \mathbf{X} \) through the operation \( \mathbf{L} \circ (\mathbf{X} \mathbf{M} \mathbf{X}^\top) \mathbf{X} \), where \( \mathbf{M} \) and \( \mathbf{L} \) are fixed matrices determined by the layer’s weights, \( \mathbf{L} \) is lower triangular, and $\circ$ denotes the Hadamard product. This formulation can be viewed as an attention mechanism where \( \mathbf{L} \circ (\mathbf{X} \mathbf{M} \mathbf{X}^\top) \) serves as the attention matrix. In this matrix, the entry at position \( (i, j) \) quantifies the degree to which the \( i \)th token attends to the \( j \)th token, analogous to the attention scores in traditional Transformer-based models.

\subsection{Individual Feature Knockout in SSMs}\label{sec:method:feature_knockout}
In both Mamba-1 and Mamba-2, each feature is treated as an independent time-varying signal modeled by distinct SSMs. Leveraging this property, we propose a feature knockout mechanism that targets specific types of features at each layer. Specifically, we classify features as either \textit{context-dependent} or \textit{context-independent} based on their decay characteristics and perform knockout by zeroing out their outputs. Formally, if we denote \(\alpha, \bar{A}\) and \(\Delta(t)\) as in \cref{seq:selective}, with \(\alpha\) being the model weights and \(\Delta(t)\) being individual token-dependent features, the dynamics of the state matrix becomes:
\[
A(t) = \bar{A}^{\Delta(t)}, \quad \prod_{t=i}^j A(t) = \bar{A}^{\sum_{t=i}^{j}\Delta(t)}.
\]
We interpret \(\prod_{t=i}^j A(t)\) as a measure of how past signals influence current representations. This product exhibits exponential decay, dependent on the matrix \(\bar{A}\) so features parameterized by \(\bar{A} \approx 0\) quickly lose memory and concentrate on localized information. In contrast, features parameterized by \(\bar{A} \approx 1\) retain more historical context. We therefore define context-dependent features as those with the largest one-third of \(\|\bar{A}\|_{1}\) values, and context-independent features as those with the lowest one-third of \(\|\bar{A}\|_{1}\) values.


\begin{figure*}[t]
    \includegraphics[width=1\linewidth]{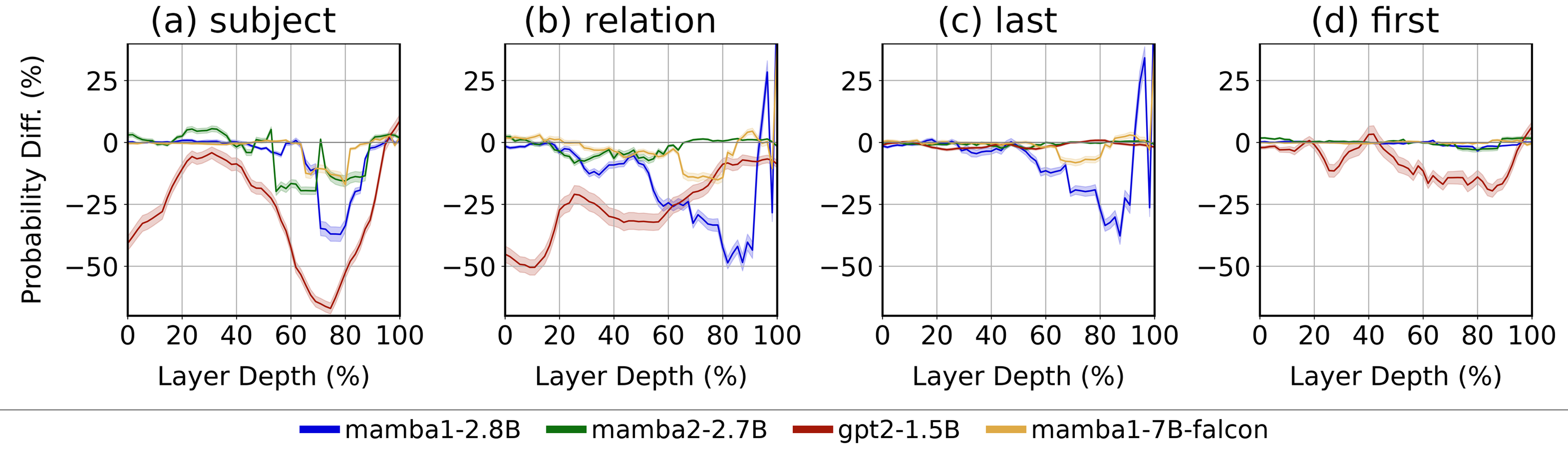}
    \caption{
    Relative change in correct-token prediction probability when removing information flow to the last token from various source tokens. The x-axis represents the relative depth of the first layer within the 9-layer attention knockout window, while the y-axis indicates the resulting performance change. The plots present model performance when knocking out (a) the subject, (b) the relation, (c) the last token, and (d) the first token. Notice that subject token knockout yields a consistent performance drop across models and similar layer-specific effects, highlighting the crucial role of subject tokens in LLMs. In contrast, knockouts of other tokens produce less robust and more variable outcomes. For ablations on dataset selection see \cref{app:exper:dataset_comparison}. See \cref {app:implementation} for extended implementation details.
    }
    \label{fig:info_flow_to_last_token}
\end{figure*}

\begin{figure}[t]
    \centering
    \includegraphics[width=1\linewidth]
    {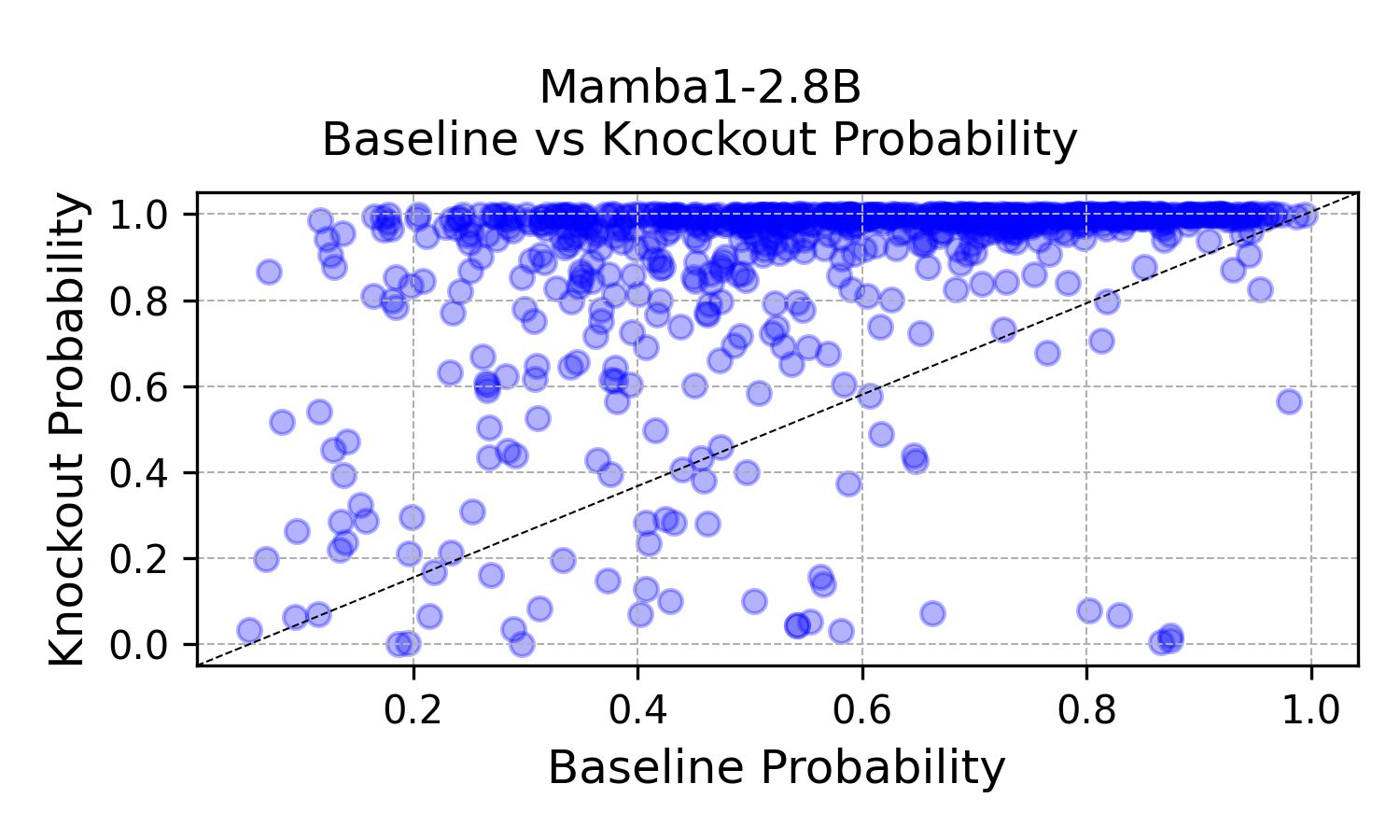}
    \caption{Impact of last token attention knockout on correct-token probability in Mamba‑1. The x-axis displays the baseline probability for each query, while the y-axis shows the probability after knockout. Observe that removing the information flow from the final token to itself across the last 9 layers yields a significant increase in correct-token probability. For further implementation details, see \cref{app:implementation}.}
    \label{fig:last_token_probabilities}
\end{figure}

\begin{figure*}[t]
    \includegraphics[width=1\linewidth]{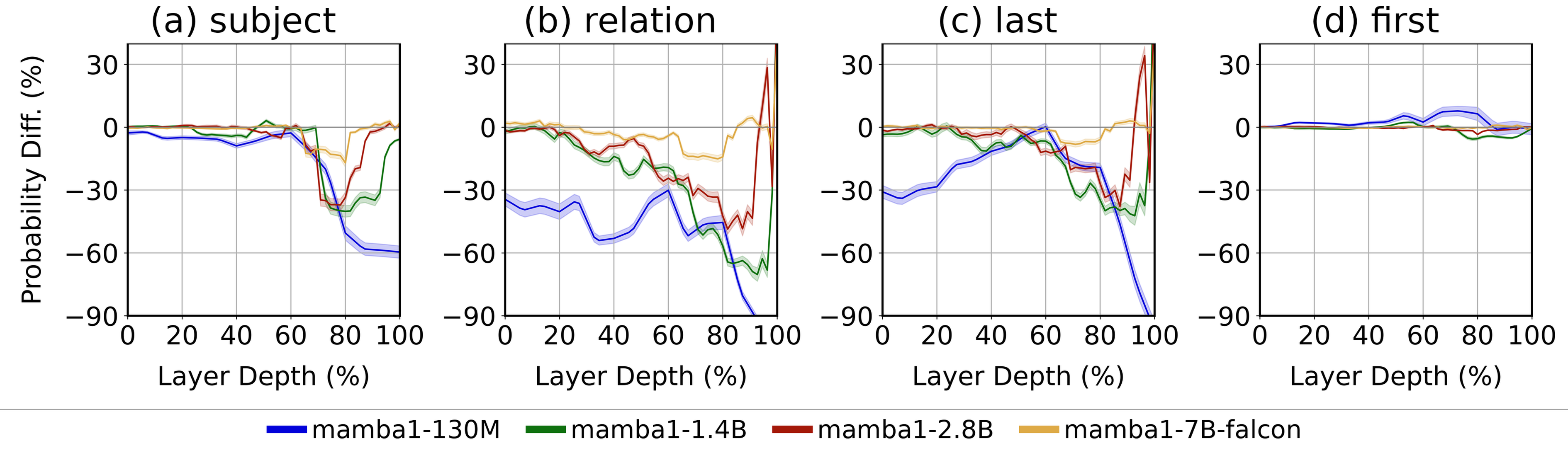}
    \caption{Relative change in correct-token prediction probability when removing information flow to the final token from various source tokens across Mamba‑1 models of varying sizes. This figure is identical to \cref{fig:info_flow_to_last_token}, but presents the performance of Mamba-1 models of sizes 130M, 1.4B, and 2.8B. As observed in other models, subject token knockout consistently reduces performance. Unique to Mamba-1, knockouts of relation tokens and of the final token in the ultimate layers initially lower the correct-token probability before a pronounced increase. For further details see \cref{fig:info_flow_to_last_token,app:implementation}. For ablations on window size see \cref{sec:exper:window_size}.}
    \label{fig:size_comparison_mamba1}
\end{figure*}

\begin{figure*}[t]
    \includegraphics[width=1\linewidth]{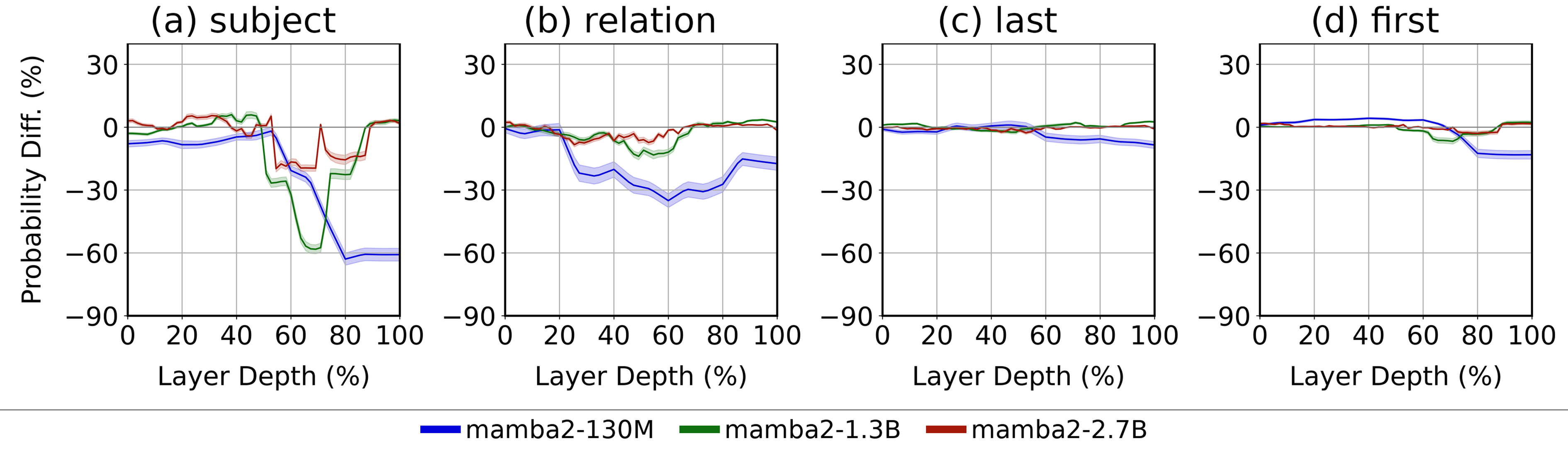}
    \caption{Relative change in correct-token prediction probability when removing information flow to the final token from various source tokens across Mamba‑2 models of varying sizes. This figure is identical to \cref{fig:info_flow_to_last_token}, but presents the performance of Mamba-2 models of sizes 130M, 1.3B, and 2.7B. As observed in other models, subject token knockout consistently reduces performance. Unique to Mamba-2, knockout of relation tokens in the later layers initially lowers the correct-token probability before a pronounced increase. For further details see \cref{fig:info_flow_to_last_token,app:implementation}. For ablations on window size see \cref{sec:exper:window_size}.}
    \label{fig:size_comparison_mamba2}
\end{figure*}

\begin{figure*}[t]
    \includegraphics[width=1\linewidth]{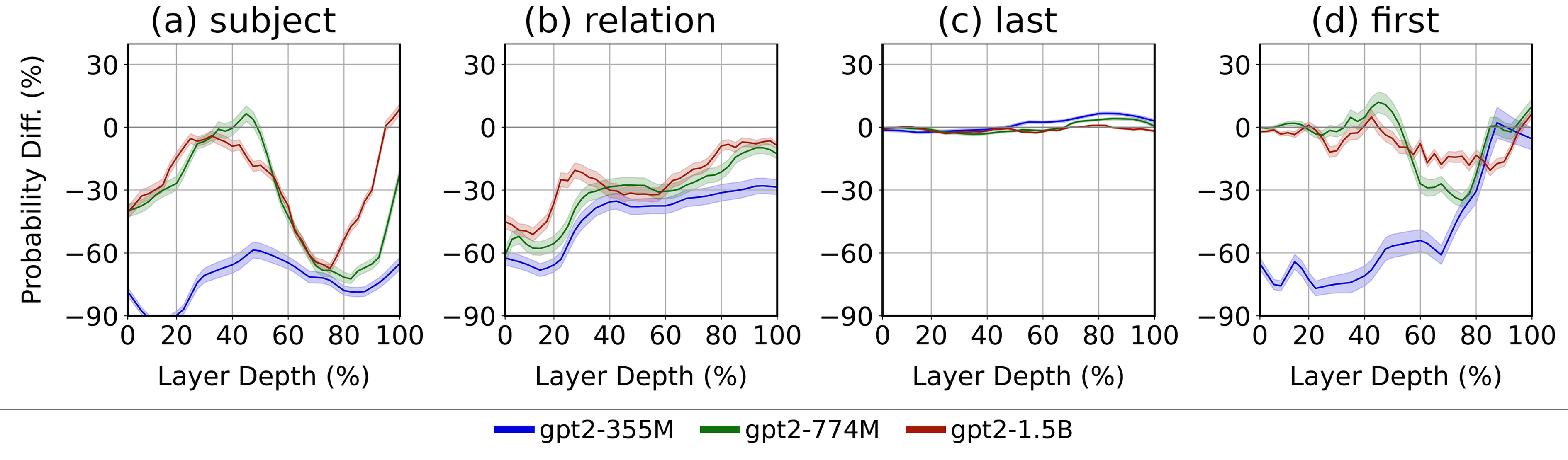}
    \captionsetup{width=1\linewidth}
    \caption{Relative change in correct-token prediction probability when removing information flow to the final token from various source tokens across GPT‑2 models of varying sizes. This figure is identical to \cref{fig:info_flow_to_last_token}, but presents the performance of GPT-2 models of sizes 355M, 774M, and 1.5B. As observed in other models, subject token knockout consistently reduces performance. Unique to GPT-2, knockout of relation tokens in earlier layers profoundly lowers the correct-token probability. Additionally, GPT-2 models exhibit a robust first token bias. For further details see \cref{fig:info_flow_to_last_token,app:implementation}.}
    \label{fig:size_comparison_gpt2}
\end{figure*}

\begin{figure*}[t]
    \centering
    \includegraphics[width=1\linewidth]{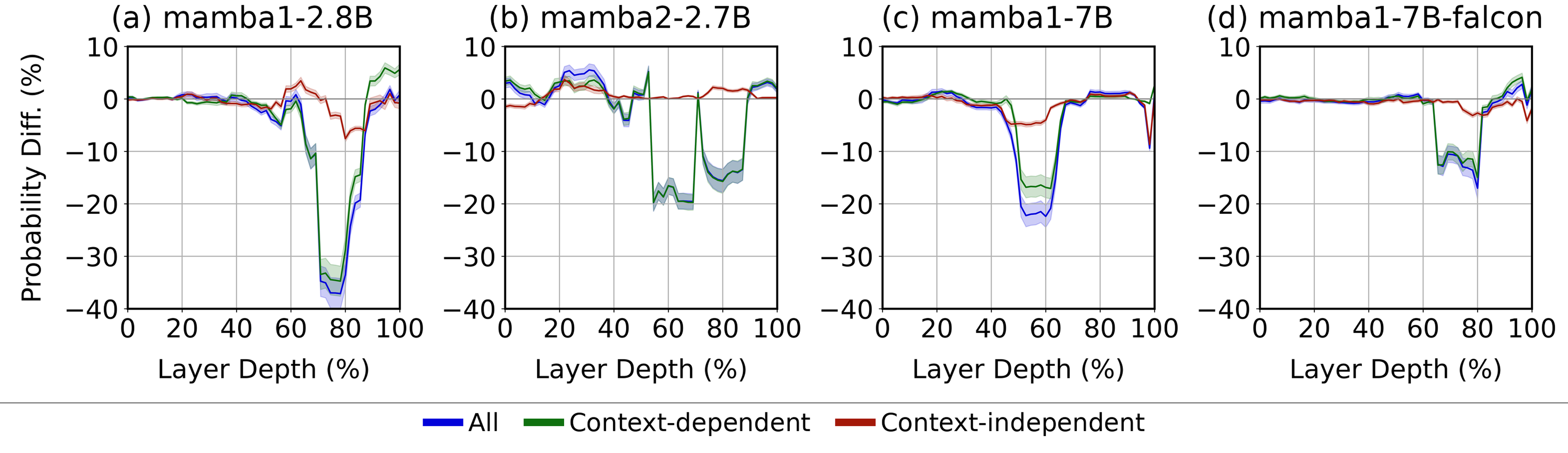}
    \caption{Relative change in correct-token probability when knocking out connections from subject tokens to the last token across different feature categories for (a) Mamba-1, (b) Mamba-2, and (c) Falcon-Mamba. We compare three settings: knockout of all features (blue), knockout of context-independent features (red), and knockout of context-dependent features (green). Notably, knocking out context-dependent features alone closely mirrors the effect of removing all features. For implementation details see \cref{app:implementation}} 
    \label{fig:feature_knockout}
\end{figure*}

\begin{figure*}
    \centering
    \includegraphics[width=1\linewidth]{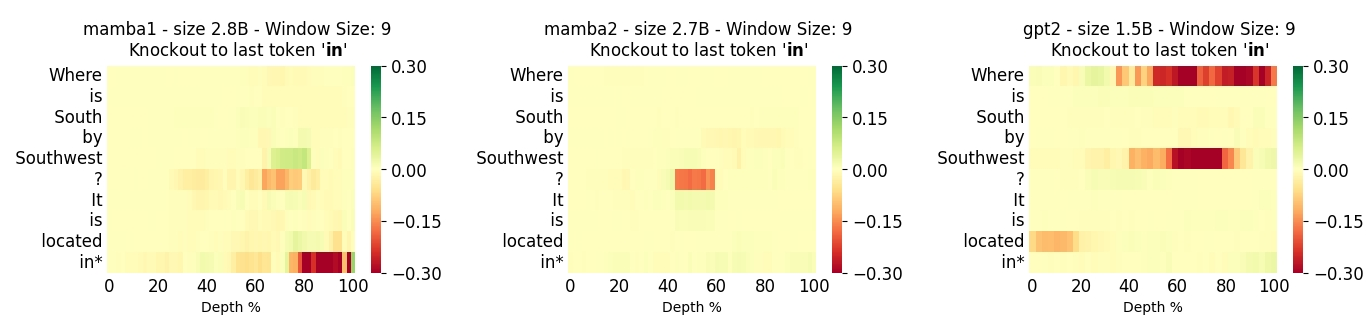}
    \caption{Relative change in correct-token probability following the removal of attention from each token to the final token. Columns indicate the first layer in the knockout window, and rows correspond to the token from which attention was removed. Results are shown for the query “Where is South by Southwest? It is located in \_\_\_”. Note that in contrast to GPT-2, neither Mamba-1 nor Mamba-2 exhibit a strong reliance on the first token, while Mamba-1 exhibits marked dependence on the final token. For additional experiments and implementation specifics, see \cref{app:exper,app:implementation}.}
    \label{fig:heatmaps}
\end{figure*}

\section{Experiments}\label{sec:exper}

This section presents our main experimental contributions and is structured as follows: \cref{sec:method:data} presents the dataset used across our experiments. \cref{sec:exper:last_tok} examines the information flow to the last token from different sources, highlighting both shared and distinct behaviors across SSM-based and Transformer-based models. \cref{sec:exper:consist} validates our findings qualitatively and shows they are robust across a variety of model sizes. \cref{sec:exper:feature_knockout} investigates the features used to convey information between tokens. In-depth case studies are explored in \cref{sec:exper:case_study}. Throughout these subsections, the results are presented with a knockout window of size \( 9 \). An analysis of the impact of window size is presented in \cref{sec:exper:window_size}, which shows that for smaller models it is better to use smaller window sizes. Other additional experiments are provided in \cref{app:exper}. Code for reproducing all experiments can be found at \url{https://github.com/nirendy/mamba-knockout}. 

\subsection{Datasets}\label{sec:method:data}
    To evaluate the performance of the models with attention knocked-out, we utilized the \texttt{COUNTERFACT} dataset \citep{meng2022locating}, which consists of factual triplets in the form (subject, relation, attribute). The task at hand requires models to predict a factual attribute (e.g., “Beats Music is owned by \_\_\_”). We focus on a subset of 672 triplets in which all models presented correctly predicted the next token. This choice disentangles the effects of the datasets and the model and ensures that any performance changes are solely due to the models. To ablate this choice, we also evaluated the effect of attention knockout on datasets composed exclusively of examples correctly predicted by each model (see \cref{app:exper:dataset_comparison}), yielding qualitatively similar results.

\subsection{Information Flow to the Last Token}\label{sec:exper:last_tok}
We investigate how the final token in Mamba-1 and Mamba-2 models transforms into the correct answer using the attention knockout method from \citet{geva2023dissecting} (see \cref{sec:method:knockout}), and compare the emerging patterns with those in the GPT-2 model \citep{radford2019language}. Specifically, we knockout the information flow to the final token from various sources—subject tokens, relation tokens, the first token, and the final token itself—with results shown in \cref{fig:info_flow_to_last_token}.

\subsubsection{Shared Characteristics}\label{sec:exper:info_flow:shared}
Across all models, inhibiting the final token’s attention to subject tokens in late-intermediate layers consistently causes a notable drop in correct-token probability, mirroring patterns observed in Transformer-based models by \citet{geva2023dissecting}. 

This result underscores the critical role of subject tokens in directing factual information flow within LLMs. More broadly, these compelling findings demonstrate that SSM-based and Transformer-based models share similar interpretability characteristics, hinting at universal properties across all attention-based architectures. In \cref{sec:exper:feature_knockout} we show a surprising connection between the behavior of subject tokens and that of context-dependent features.

\subsubsection{Distinct Characteristics}\label{sec:exper:info_flow:distinct}
In addition to their shared characteristics, our analysis reveals that each model exhibits unique properties not present in all others.

First, we observe that, consistent with \citet{geva2023dissecting}, GPT-2 exhibits a pronounced first-position bias—a feature not shared by the SSM-based models. This finding aligns with previous research suggesting that the first-position attention sink is more evident in normalization-based attention methods \citep{gu2024attentionsinkemergeslanguage}.

Second, GPT-2 shows a strong dependence on information flow from relation tokens in the very early layers, a reliance that gradually diminishes in later layers. In contrast, SSM-based models only exhibit significant dependence on relation tokens in the later layers. Notably, both Mamba-1 and Falcon-Mamba display a pronounced knockout response, characterized by an initial drop in performance followed by a sharp increase in correct-token probability.

Surprisingly, we observe that both Mamba-1 and Falcon-Mamba also exhibit a marked increase in the correct-token probability at later layers when attention from the last input token is blocked. To investigate this phenomenon, we compare the correct token probability before and after the knockout in \cref{fig:last_token_probabilities}, finding that it consistently surges to nearly $1$, regardless of the initial likelihood. We find this result surprising and defer a deeper exploration of its implications to future work.

\subsection{Architecture Consistency}\label{sec:exper:consist}
To assess whether this pattern of results should be attributed to architectural differences, we validate our findings by performing attention knockout on models of different sizes, presented in \cref{fig:size_comparison_mamba1,fig:size_comparison_mamba2,fig:size_comparison_gpt2}. Our results show that while the smallest models seem to suffer from knockout much more substantially than the larger models (which could be explained due to the fact that our window size of choice blocks 37.5\% of layers - see Section~\ref{sec:exper:window_size} for a more thorough analysis of this behavior) the overall patterns are consistent, with the main phenomenon shared throughout---all models show a reliance on direct information flow from the subject tokens to the last token in the late-intermediate layers. Conversely, as discussed in \cref{sec:exper:info_flow:distinct}, the models exhibit distinct, architecture-specific phenomena; Mamba‑1 models rely on both the last token and relation tokens in the later layers (see \cref{fig:size_comparison_mamba1}); Mamba‑2 models primarily depend on relation tokens in the later layers (see \cref{fig:size_comparison_mamba2}); GPT‑2 models show a strong dependence on relation tokens in the early layers, with this reliance gradually diminishing in later layers (see \cref{fig:size_comparison_gpt2}). To reinforce the claim that the unique information flow patterns in Mamba variants arise from the SSM layers and their characteristics, we extended our information flow analysis to two additional variants of transformer-based LLMs, Llama 3 \citep{grattafiori2024llama} and Mistral \citep{jiang2024identifying}. The results show a consistent pattern shared between all transformer-based LLMs which is distinct from the patterns observed in Mamba-based models. \Cref{app:exper:transformer} presents detailed experiments on this.

\begin{figure*}[t]
    \centering
    \includegraphics[width=1\linewidth]{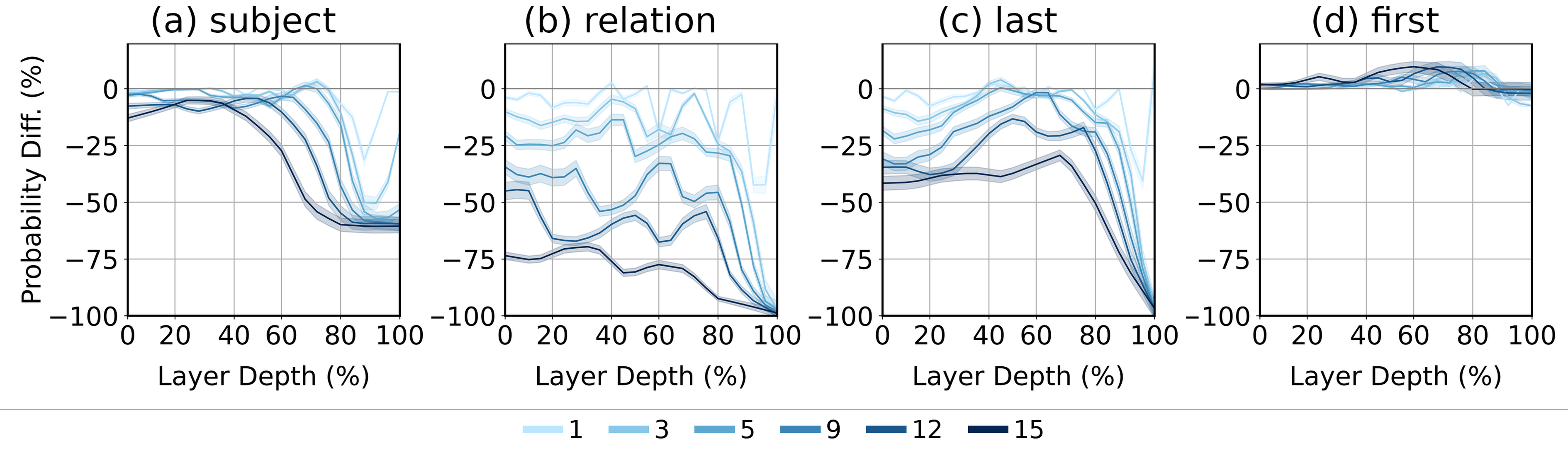}
        \caption{Relative change in correct-token probability after removing the information flow to the final token from various source tokens for Mamba‑1 130M across different window sizes. This figure, which parallels \cref{fig:info_flow_to_last_token}, displays performance for window sizes of $1$, $3$, $5$, $9$, $12$, and $15$. The qualitative patterns observed for Mamba-1 remain consistent, with larger window sizes amplifying these effects. For additional details, see \cref{fig:info_flow_to_last_token,app:implementation}.}
    \label{fig:ws_mamba1_130M}
\end{figure*}

\begin{figure*}[t]
    \centering
    \includegraphics[width=1\linewidth]{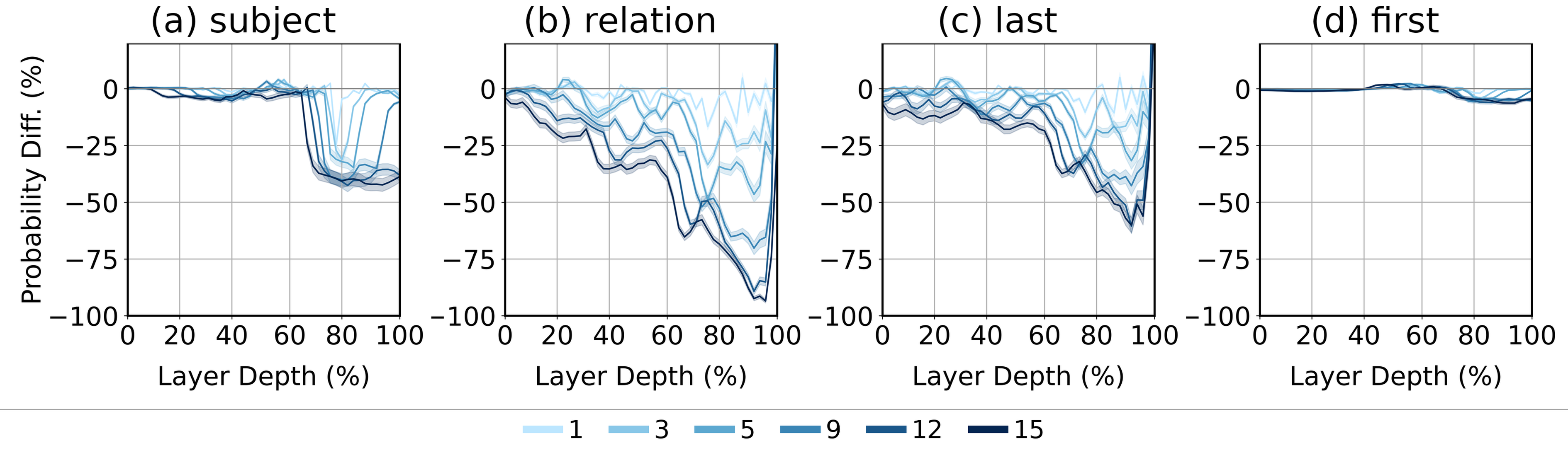}
        \caption{Relative change in correct-token probability after removing the information flow to the final token from various source tokens for Mamba‑1 1.4B across different window sizes. This figure is identical to \cref{fig:ws_mamba1_130M} but instead presents the results for Mamba-1 1.4B. For additional details see \cref{fig:ws_mamba1_130M,fig:info_flow_to_last_token,app:implementation}.}
    \label{fig:ws_mamba1_1.4B}
\end{figure*}

\begin{figure*}[t]
    \centering
    \includegraphics[width=1\linewidth]{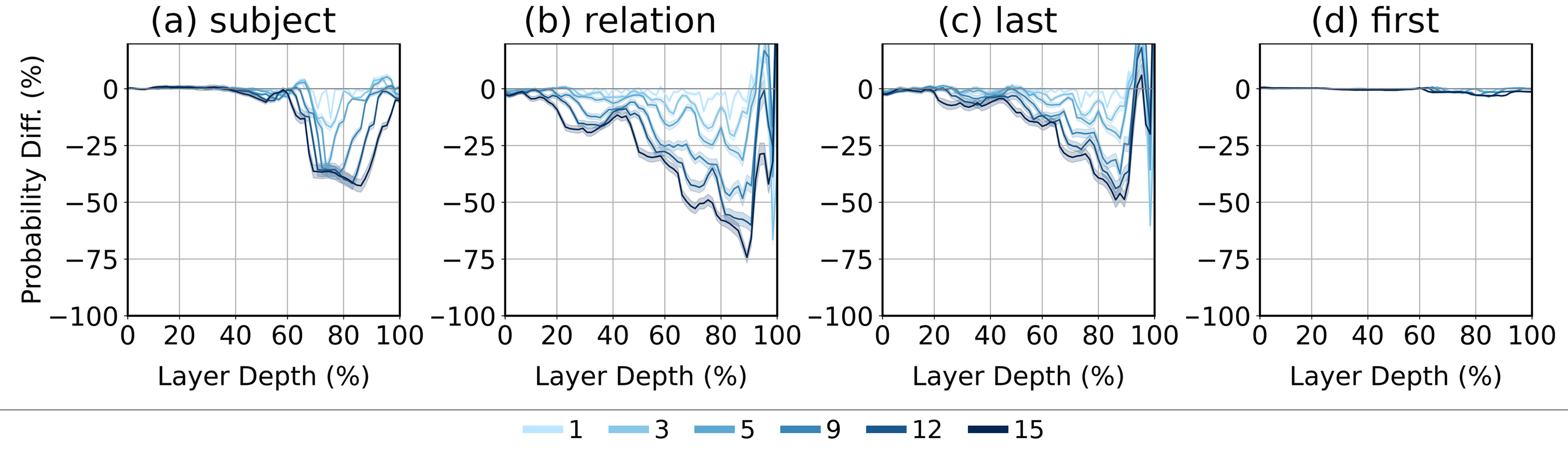}
        \caption{Relative change in correct-token probability after removing the information flow to the final token from various source tokens for Mamba‑1 2.8B across different window sizes. This figure is identical to \cref{fig:ws_mamba1_130M} but instead presents the results for Mamba-1 2.8B. For additional details, see \cref{fig:ws_mamba1_130M,fig:info_flow_to_last_token,app:implementation}.}
    \label{fig:ws_mamba1_2.8B}
\end{figure*}

\subsection{Feature Knockout}\label{sec:exper:feature_knockout}
Expanding on \cref{sec:method:feature_knockout}, we selectively perform attention knockout on context-dependent and context-independent features to investigate their distinct roles. We compare feature knockout from subject tokens to the last token under three conditions: (1) knocking out \emph{all} features (as in \cref{sec:exper:last_tok}), (2) knocking out only \emph{context-dependent} features, and (3) knocking out only \emph{context-independent} features. \cref{fig:feature_knockout} shows that performing knockout on context-dependent features (slower decay) yields behavior similar to performing attention knockout in all features. In contrast, performing knockout in context-independent features (faster decay) shows little to no impact. These observations underscore the pivotal role of context-dependent features in subject tokens, effectively mirroring the impact of knocking out all features. While this analysis remains correlational and does not yet establish causation, it opens promising directions for future interpretability research in SSM-based architectures: much like how attention maps offer interpretability insights in Transformer architectures, evaluating a feature’s degree of context dependence in SSM-based models could provide a valuable lens into how tokens behave across layers and influence the model’s decision making. Furthermore, building on these insights, future work could potentially investigate whether selectively fine-tuning context-dependent features alone enables more efficient training in SSM-based architectures. \cref{app:practical} proposes several practical applications, while a full exploration of their potential is left to future investigations. We extended our analysis of context-dependent and independent features to additional variants of Mamba-1 and Mamba-2 models in \cref{app:exper:feature_knockout}. The supplementary findings reaffirm our conclusions.

\subsection{Specific Case Studies}\label{sec:exper:case_study}
Our final contribution is a detailed analysis of specific examples and the impact of attention knockout from different tokens on model performance. An example is shown in \cref{fig:heatmaps}. Our experiments reveal two key findings. First, Mamba-1 and Mamba-2 models do not exhibit a profound first-token bias seen in Transformer-based models \citep{xiao2024efficientstreaminglanguagemodels}, as blocking the first token from attending to the last token does not significantly affect performance.
Second, Mamba-1 models heavily rely on the last token, a behavior not observed in Mamba-2 models as can also be seen in \cref{fig:info_flow_to_last_token}. We attribute this difference to architectural variations between the models and anticipate that future work will explore its underlying causes. More examples are available in \cref{app:exper:heatmaps}.


\subsection{Effect of Window Size}\label{sec:exper:window_size}
The choice of window size for the knockout analyses can produce varied results. On one hand, choosing too small a window may show little to no effect, as information flow could be distributed across multiple consecutive layers. On the other hand, choosing too large a window reduces resolution, and impairs our ability to reach meaningful conclusions regarding information flow in the model. 

To empirically examine the effect of window size, we perform the knockout experiment in \cref{sec:exper:last_tok} using several window sizes - $1$, $5$, $9$, $12$, $15$, on both Mamba-1 (\cref{fig:ws_mamba1_130M}, \cref{fig:ws_mamba1_1.4B}, \cref{fig:ws_mamba1_2.8B}) and Mamba-2 (\cref{fig:window_size_ablation_mamba2} in \cref{app:exper}) of various sizes. Results corroborate our intuition, with reduced window size showing lesser effects, and higher window sizes showing greater effects. Larger window sizes display smoother, more diffuse patterns, while smaller window sizes are characterized by short surges, indicating specific layers are more critical in the flow of information. 
While the overall pattern of information flow remains the same, with the critical point of subject-to-last token information flow in the late-intermediate layers remaining the same, we can see that it is better to use smaller window sizes with smaller models. While in ~\cref{fig:size_comparison_mamba1,fig:size_comparison_mamba2,fig:size_comparison_gpt2} we see a different behavior  for the small models, if we look at their behavior with smaller window sizes, it is more similar to the larger models. The reason for that is that the smaller model has less parameters and, therefore, larger window sizes blocks a large portion of the information, e.g., blocking 37.5\% of blocks for window size 9. Therefore, it might be better to use a smaller window size with knockout to get a better resolution with small models and in this way better analyze their behavior.

\section{Conclusion}\label{sec:conclusion}
We observe some similarities between SSM-based and Transformer-based models in extracting factual attributes and transferring information from the subject to the last token. We show that the models rely on different parts of the sentence to extract the correct answer, depending on their architecture. Using a unique property of SSM-based models, we identify the functionality of specific features based on their weights.

This work serves as both an initial exploration of the unique properties of SSM-based language models and a deeper investigation of the shared foundations of language understanding across neural network models. The explainability approach presented here may lead to more unified methods for explaining the internal computations of diverse neural architectures. 

Our findings offer several key contributions. They can guide training decisions for language models by enabling targeted pruning that eliminates redundant information flows and reduces computational overhead. Moreover, our work lays the groundwork for fine-tuning strategies—such as adjusting Mamba's final layer—to modify language distribution without sacrificing factual accuracy. Finally, distinguishing between shared and unique patterns across architectures can deepen our understanding of language modelling as a whole, and inform the design of more robust AI models.

\section{Limitations}
In this work we primarily use attention-knockout to examine how information flows between tokens in Mamba-1 and Mamba-2. While this approach shows causality, when severing token connections changes the correct attribute probability, it is far from exhausting the complete pattern of information flow in the model. First, our use of this approach is limited by its application in continuous layer windows, with all SSMs in a layer affected simultaneously. A much more diffuse and intricate pattern could occur in reality, while this approach would fail to recognize it. For a more holistic analysis, fine-tuning-based approaches should be used.
While we identify critical connections between tokens, our method does not elaborate on what information actually passes through these connections. Future works should attempt to decipher the content of these internal representations.
As the attention knockout effect is not ecological, any change in probability could be attributed to it, however the localized pattern suggests that changes are due to specific missing information.
Attention-knockout focuses only on the flow of information between tokens, thus it cannot be used in order to explain the role of token-independent operations (e.g. gating, convolution, etc.). As stated in \cref{sec:related}, we focus on between-token information flow as it is the key difference between selective-SSMs and attention.
Our methods recognize how information flows between tokens and the critical points in this contextualization, however, they do not shed light on why both Transformers and Mamba models converge to these similar patterns. Future works should attempt to explain the inductive-biases of the different architectures that drive these differences in information flow.

\section*{Acknowledgements}
IDG, YRM and YS are supported in part by the Tel Aviv University Center for AI and Data Science.


\bibliography{custom}

\appendix

\crefalias{section}{appendix}
\crefalias{subsection}{appendix}
\crefalias{subsubsection}{appendix}

\section{Extended Related Work}
\label{app:related_extended}

Our work contributes to the growing field of mechanistic interpretability, which seeks to explain neural network behaviors by analyzing internal circuits. The circuit analysis paradigm was introduced by \citet{olah2020zoom}, who proposed that groups of neurons form interpretable sub-networks (circuits). \citet{elhage2021mathematical} extended this approach to Transformers by reverse-engineering a small GPT-2 model, identifying induction-head circuits that facilitate in-context copying. Similarly, \citet{olsson2022context} explored how certain attention heads enable in-context learning.

Beyond attention mechanisms, other research examined MLP pathways and feature representations. \citet{geva2021transformerfeedforwardlayerskeyvalue} demonstrated that Transformer MLP layers can act as key-value memories for factual associations \citep{meng2022locating}. Complementary studies by \citet{finlayson2021causal} used causal interventions to trace linguistic features through Transformer layers, revealing assemblies of neurons mediating specific features. Recent work on polysemantic versus monosemantic neurons \citep{elhage2022toy,bricken2023towards} further highlights the importance of disentangling neuron functions to improve interpretability.

Our Mamba-based approach for intervening in information flow aligns with prior methods of causal tracing and circuit testing. \citet{meng2022locating,meng2023masseditingmemorytransformer} introduced causal tracing to identify critical layers for factual storage, using targeted edits (ROME and MEMIT) for knowledge modification. \citet{hase2023doeslocalizationinformediting} similarly probed Transformer layers to localize factual information. Our attention-knockout method builds upon these studies, adapting fine-grained intervention techniques to the state-space architecture of Mamba.

We also connect to automated circuit discovery techniques such as path patching \citep{wang2022interpretability,conmy2023towards}, where researchers systematically test causal paths within models. Our knockout method similarly isolates token-to-token information channels, providing insight into causal computation pathways.

\section{Practical Applications}
\label{app:practical}

The interpretability methods presented in this work provide practical directions for targeted model interventions. We propose a few of them below.

\textbf{Pruning.} Our distinction between context-dependent features—those critical for inter-token information flow—and context-independent features—focused primarily on single-token processing—enables smarter pruning strategies. Context-independent features often contribute minimally to between-token information-flow and therefore can be safely pruned, reducing model complexity without significant performance loss. This targeted approach mirrors prior work in Transformers where redundant attention heads were pruned effectively \citep{michel2019sixteen}.

\textbf{Fine-Tuning and Adaptation.} Context-dependent features identified via knockout analysis represent targeted fine-tuning opportunities. Adjusting only these crucial features could efficiently adapt the model to new tasks or factual domains without global retraining. This focused fine-tuning strategy aligns with prior findings in Transformers \citep{hase2023doeslocalizationinformediting} where precise adjustments to identified causal pathways effectively updated model knowledge. Our methods similarly offer practical leverage points for precise and efficient model adaptation.

\textbf{Model Editing.} Token-level knockout identifies key token interactions critical for factual recall, highlighting optimal sites for targeted model edits. While similar causal tracing in Transformer models has informed successful editing strategies \citep{meng2022locating,meng2023masseditingmemorytransformer}, our approach extends these methods to SSM-based models such as Mamba-1 and Mamba-2. In summary, the interpretability techniques introduced here serve as diagnostic tools to inform strategic model modifications, pruning, and fine-tuning.

\section{Further Experiments}\label{app:exper}

\subsection{Dataset Ablations}
\label{app:exper:dataset_comparison}

\begin{figure*}[t]
    \centering
    \includegraphics[width=1\linewidth]{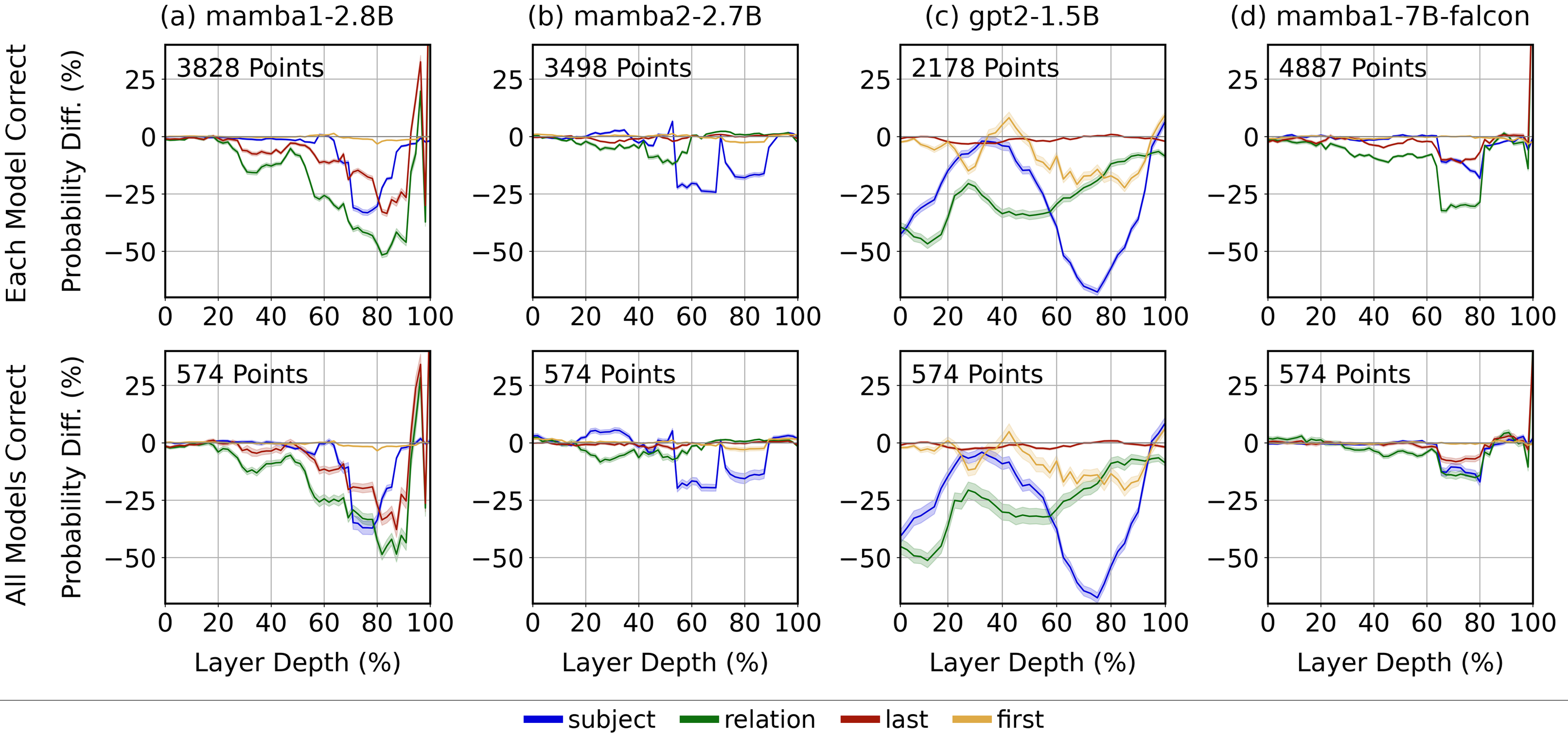}
    \caption{Relative change in correct-token probability after knocking out the connection from each token to the last token in (a) Mamba-1 2.8B, (b) Mamba-2 2.7B, (c) GPT-2 1.5B, (d) Mamba-1 7B Falcon. The top row shows results on the subset of samples each model was correct on. The bottom row shows results on the subset of samples where all models were correct. The X-axis indicates the relative depth of the layer on which we perform the knockout. The y-axis indicates the relative change in correct-token probability. Different colors indicate the source token.}
    \label{app:fig:ds_ablate}
\end{figure*}

We analyzed a subset of factual associations where all models correctly predicted the next token to disentangle the effects of the datasets and the model and ensure that any performance changes are solely due to the models. A potential drawback of this approach is that the selected subset could introduce bias, potentially leading to misleading conclusions. To assess whether this is the case, we performed the same knockout analysis on the set of sentences for which each model made a correct prediction, allowing for a broader comparison. The results of these evaluations are shown in \cref{app:fig:ds_ablate}. In addition, we present results on the subset of inputs for which all models correctly predicted the next token. The figure shows that the same patterns emerge across these subsets with minimal differences, confirming that our approach robustly identifies consistent phenomena across the different models.

\subsection{Further Comparisons with Transformer-Based Models}
\label{app:exper:feature_knockout}

\begin{figure*}[t]
    \centering
    \includegraphics[width=1\linewidth]{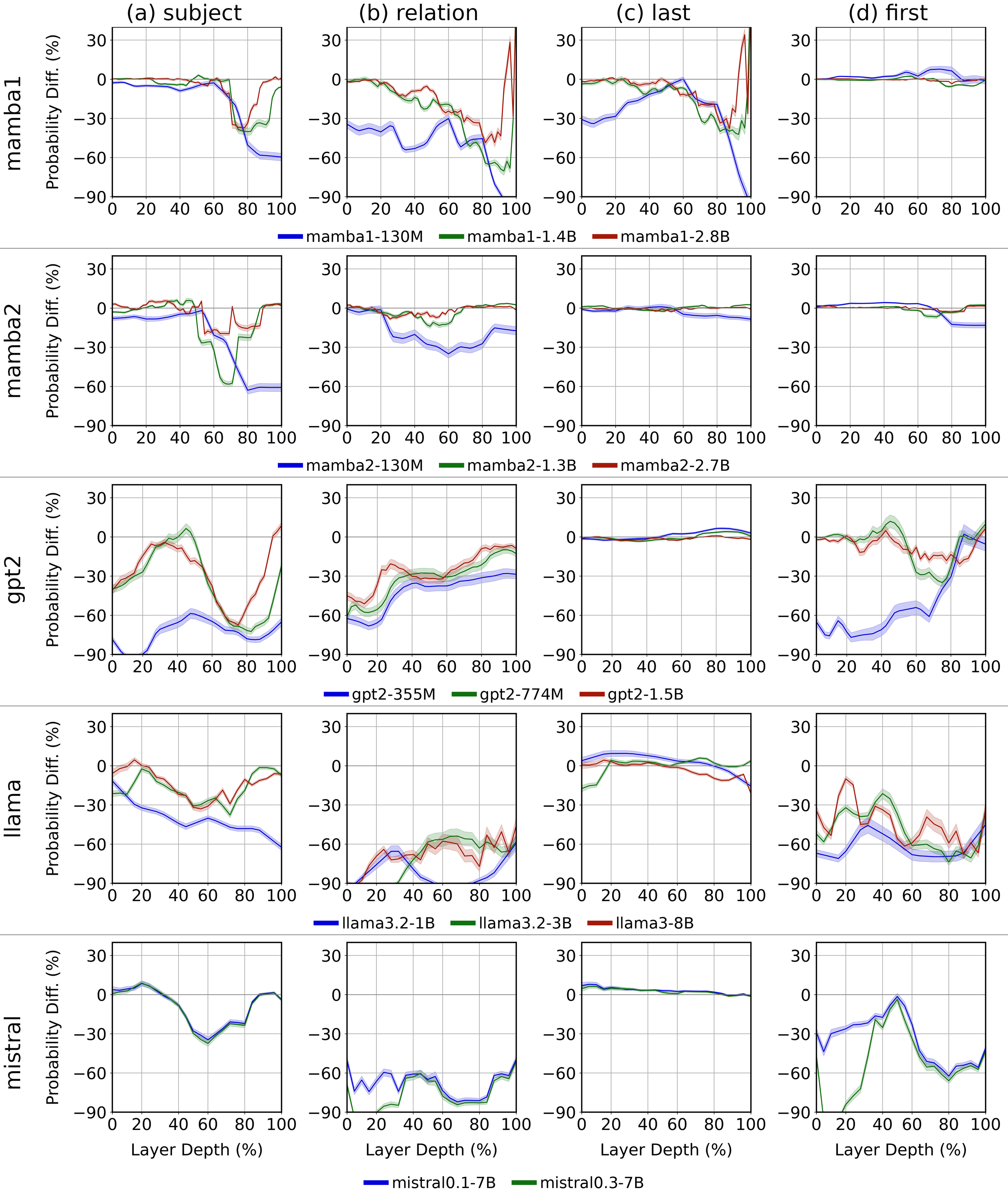}
    \caption{A full comparison of the effect of attention knockout on all model variants. Each column indicates the knockout target. Each row indicates the model family (Mamba-1, Mamba-2, Falcon Mamba, GPT-2, Llama, Mistral). Different coloring indicates the specific model variant.}
    \label{fig:app:arch}
\end{figure*}

To confirm that the difference in information flow patterns between Mamba models \citep{gu2023mamba,dao2024transformers} and GPT-2 models \citep{radford2019language} can be attributed to architectural differences, we report auxiliary information flow experiments performed on additional Transformer-based LLMs. Specifically, we analyzed three variants of the Llama 3 model \citep{grattafiori2024llama} and two variants of the Mistral model \citep{jiang2024identifying}. The results are reported in \cref{fig:app:arch}. All transformer-based variants show shared characteristics, implying they should be attributed to the shared architecture: \textit{(i)} all models show a similar pattern of a critical flow of information passing from the subject tokens to the last token in the late intermediate layers; \textit{(ii)} all models give little importance to information flow from the last token to itself; and; \textit{(iii)} nearly all models show strong evidence for first-position bias, similar to previous findings \citep{xiao2024efficientstreaminglanguagemodels}, with the most substantial impact observed in the final layers.

\subsection{Feature Knockout on Additional Models}
\label{app:exper:transformer}

\begin{figure*}[t]
    \centering
    \includegraphics[width=0.9\linewidth]{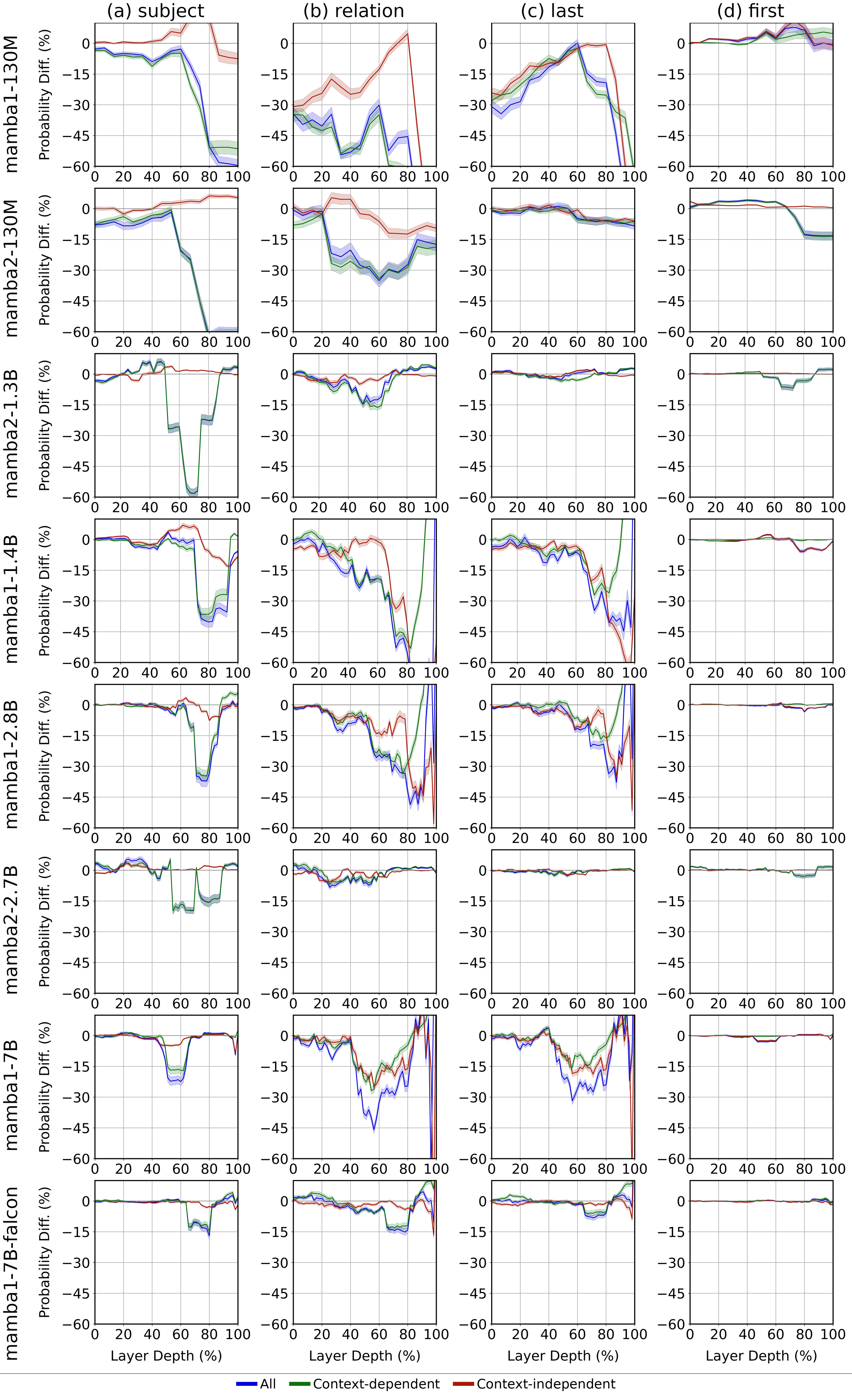}
    \caption{A full comparison of the effect of feature-knockout on all size variants of Mamba models. Each column shows the effect when knocking out information flow from (a) the subject, (b) the relation, (c) the last, and (d) the first tokens to the last token. Each row shows the results for a different model.}
    \label{fig:app:feature_knockout}
\end{figure*}

To further examine the role of context-dependent features in Mamba models, we performed the same feature-knockout analysis on smaller variants of the Mamba-1 and Mamba-2 models. Results are reported in \cref{fig:app:feature_knockout}. All models show consistent roles for context-dependent and independent features, with knockout of context-dependent features from subject tokens having a much greater impact than that of context-independent features. On the other hand, knockout of context-dependent features from either the relation tokens, the last token or the first token shows similar patterns to knockout of context-independent features.

\subsection{Specific Case Studies}
\label{app:exper:heatmaps}

    We extend the analysis of \cref{sec:exper:case_study} to additional examples, architectures, and knockout-window sizes.

    \paragraph{Across Architectures and Model Sizes.}
    
\begin{figure*}[t]
    \centering
    \includegraphics[width=1\linewidth]{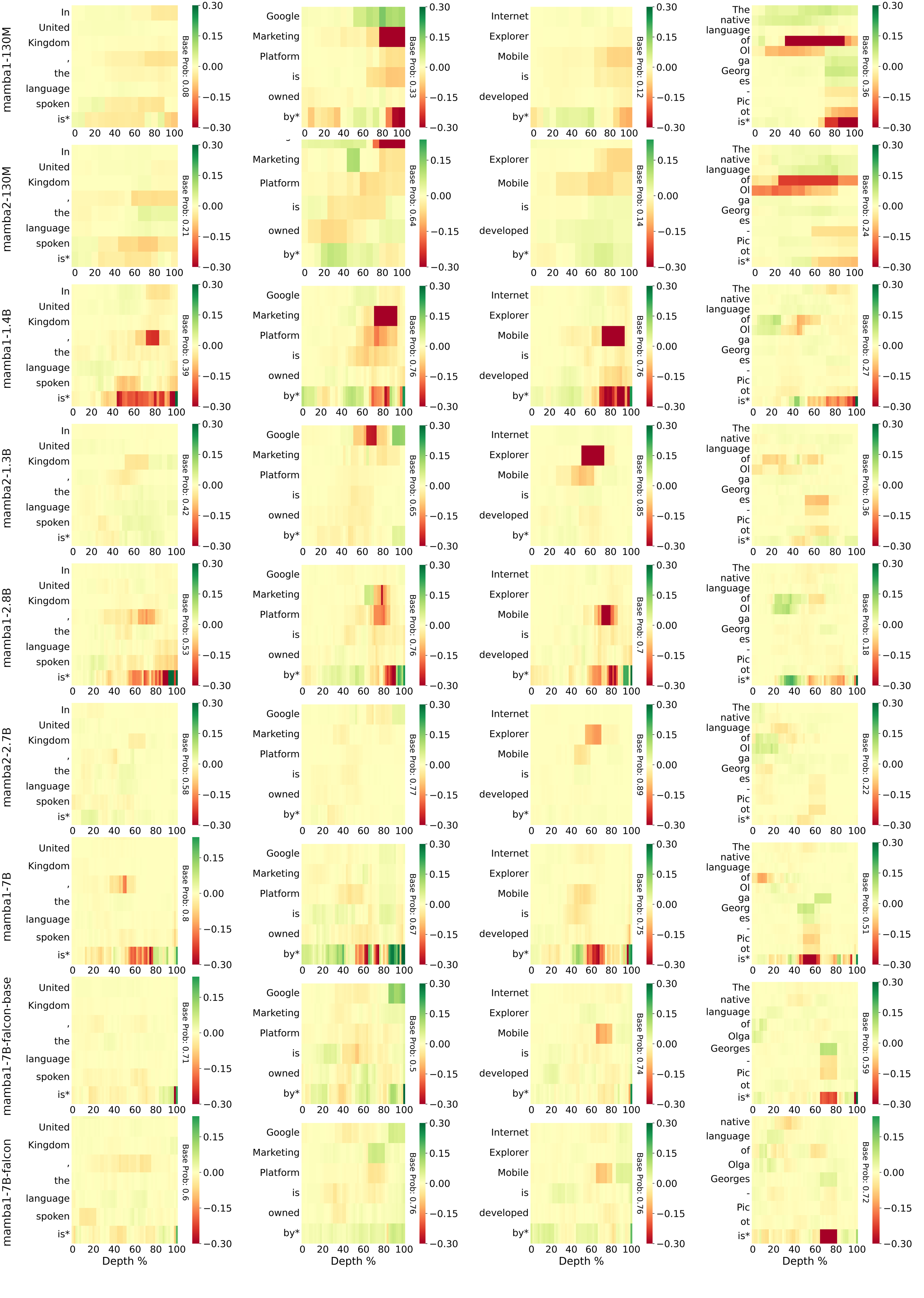}
        \caption{Relative change in correct-token probability after ablating the connection from each token to the final token, shown across different Mamba architectures with a fixed window size of 9. While Mamba-1 shows strong reliance on the final token and minimal dependence on early tokens, Mamba-2 distributes influence more evenly, suggesting improved integration of contextual information.}
    \label{fig:heatmaps_per_model_sizes_and_archs_mamba}
\end{figure*}

\begin{figure*}[t]
    \centering
    \includegraphics[width=1\linewidth]{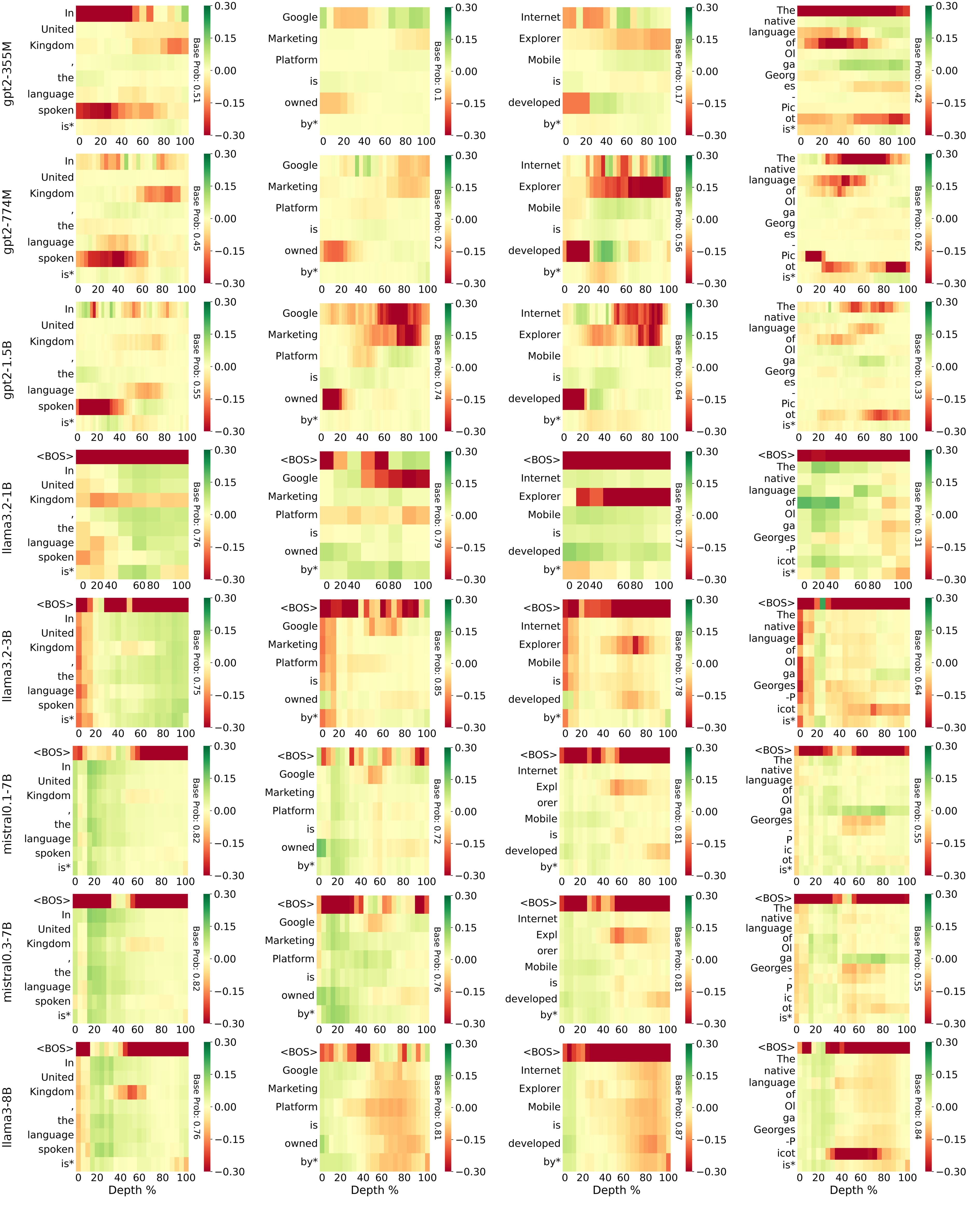}
        \caption{Same analysis as above (\cref{fig:heatmaps_per_model_sizes_and_archs_mamba}), but applied to Transformer architectures at comparable parameter scales, using a fixed window size of 9.}
    \label{fig:heatmaps_per_model_sizes_and_archs_transformer}
\end{figure*}

    \Cref{fig:heatmaps_per_model_sizes_and_archs_mamba} displays ablation heatmaps for Mamba-1 and Mamba-2.  The patterns highlighted persist: Mamba-1 shows a pronounced reliance on the final token with little first-token bias, whereas Mamba-2 distributes influence more evenly across the sequence. In both Mamba variants, we also observe a concentrated band of sensitivity around the subject tokens - the same band that emerged in the information-flow graphs at roughly the 70 depth layer—underscoring the models’ focus on the entity whose attribute is being predicted. The same analysis for Transformer baselines is presented in \cref{fig:heatmaps_per_model_sizes_and_archs_transformer}, where a strong first-token dependence is evident at every size.
    
    \paragraph{Effect of Knockout-Window Size.}  

\begin{figure*}[t]
    \centering
    \includegraphics[width=1\linewidth]{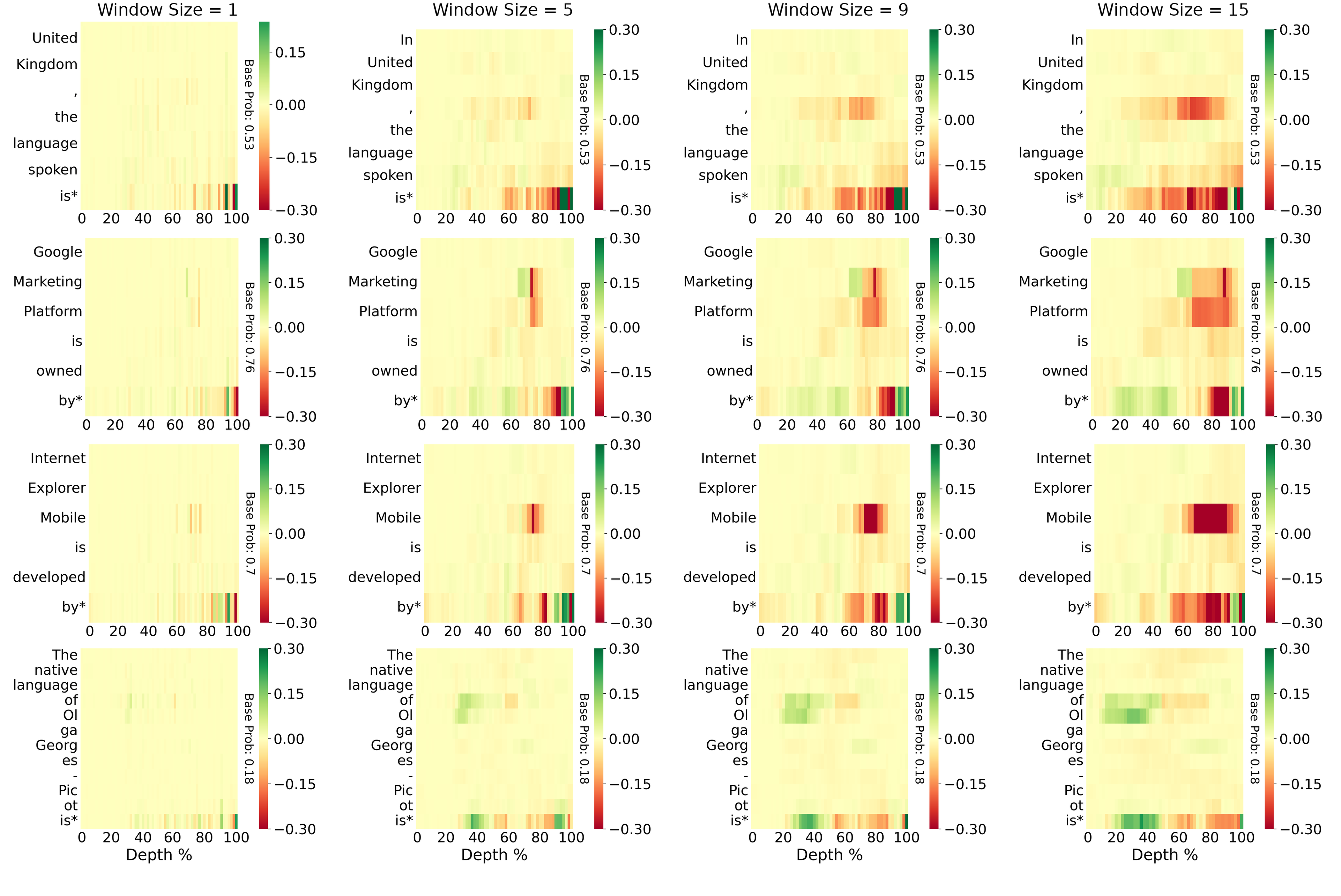}
        \caption{Relative change in correct-token probability after ablating the connection from each source token to the last token in Mamba-1 2.8 B. Columns correspond to ablation window sizes $\in\{1,5,9,15\}$, and rows index the source token. Enlarging $W$ amplifies the impact of mid-sequence tokens.
}
    \label{fig:heatmaps_window_size_mamba1-2.8B}
\end{figure*}

\begin{figure*}[t]
    \centering
    \includegraphics[width=1\linewidth]{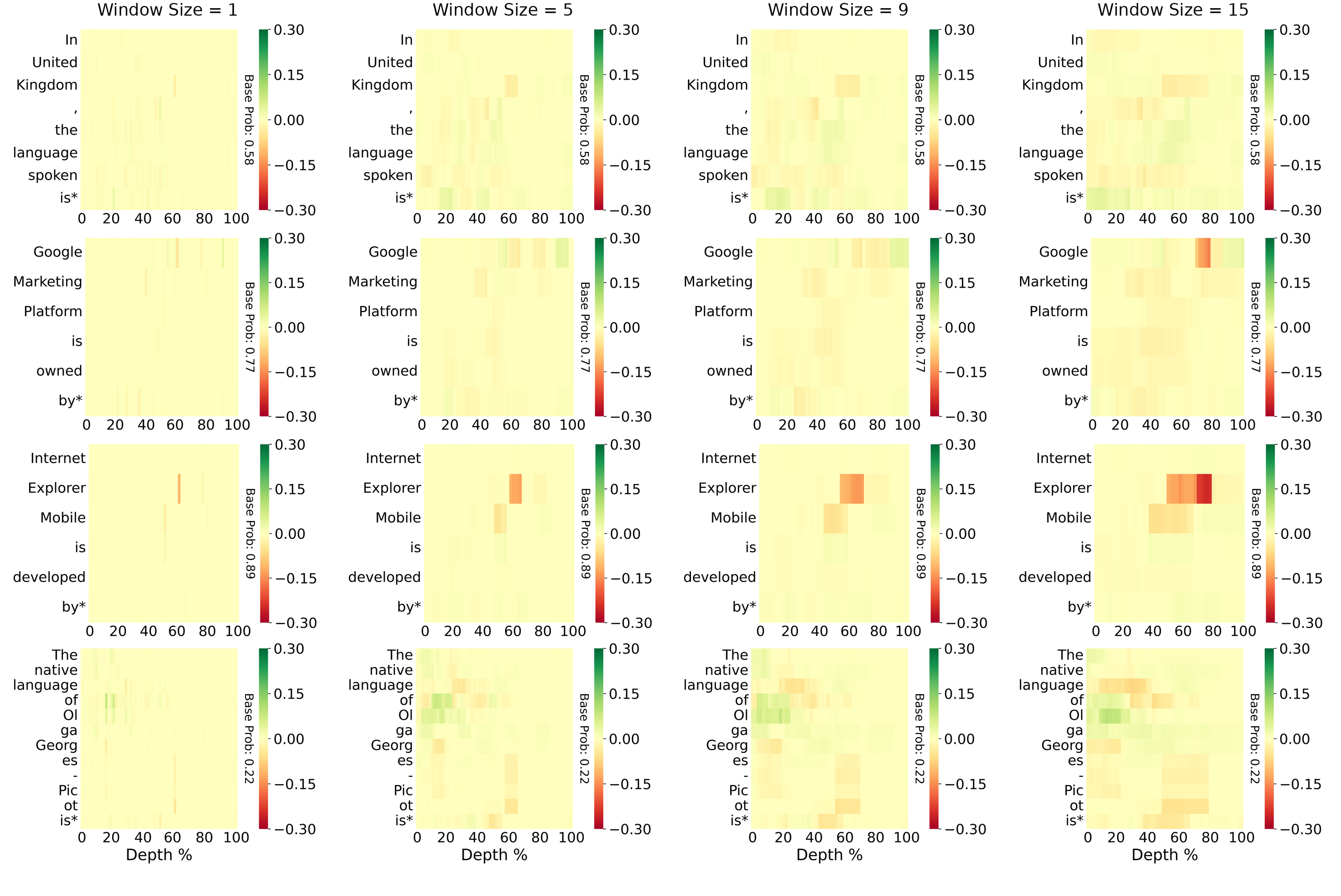}
        \caption{Identical analysis for Mamba-2 2.7 B. Columns again denote ablation window sizes $\in\{1,5,9,15\}$. Relative influence is more evenly distributed than in Mamba-1, with larger $W$ revealing a cumulative contribution from earlier context.
}
    \label{fig:heatmaps_window_size_mamba2-2.7B}
\end{figure*}

\begin{figure*}[t]
    \centering
    \includegraphics[width=1\linewidth]{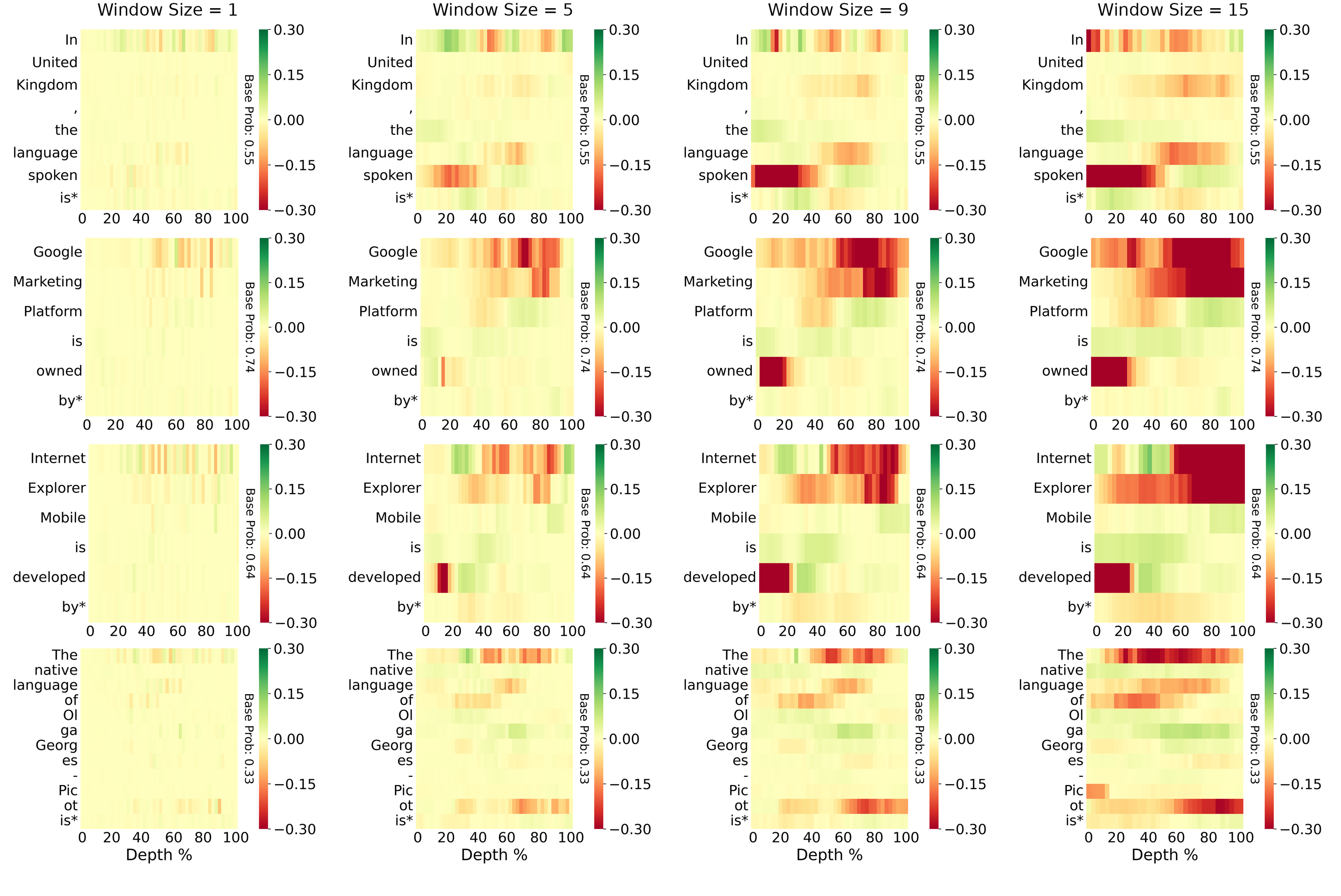}
        \caption{Same analysis applied to GPT-2 1.5 B. Columns show ablation window sizes $\in\{1,5,9,15\}$.}
    \label{fig:heatmaps_window_size_gpt2-1.5B}
\end{figure*}

    To isolate how window size modulates these effects, we vary $WS\in{1,5,9,15}$ in the selected model architectures and sizes (\cref{fig:heatmaps_window_size_mamba1-2.8B,fig:heatmaps_window_size_mamba2-2.7B,fig:heatmaps_window_size_gpt2-1.5B}).
    Smaller $WS$ values produce sparser heatmaps, but pinpoint the layers that drive the largest probability shifts.  For example, the surge in final-token probability for Mamba-1 can be traced to ablation in the single last layer ($WS=1$).  Similar layer-specific contributions are visible for individual phrases in both Mamba-2 and GPT-2.  In contrast, some effects only emerge when the ablation window is sufficiently deep: in GPT-2, the characteristic dependence of the first token is barely detectable at $WS=1$ but becomes more pronounced as the window expands

\section{Implementation Details}
\label{app:implementation}

Code for reproducing all experiments can be found at \url{https://github.com/nirendy/mamba-knockout}. All experiments were conducted on a single NVIDIA A100 GPU and implemented using PyTorch \citep{paszke2019pytorchimperativestylehighperformance}. The attention-based Mamba-2 model was evaluated within the same environment across all experiments to ensure consistency.
For the Mamba-1 analyses, we used pretrained weights from \citet{gu2024mambalineartimesequencemodeling} published in the Hugging Face framework \citep{wolf2020huggingfacestransformersstateoftheartnatural}.
For the Mamba-2 analyses, we modified the implementation of \citet{mamba2-github}. As the original Mamba-2 implementation doesn't compute the attention matrix, relying instead on a more efficient computation strategy, we re-implemented the selective SSM layer to explicitly construct the attention matrix while maintaining the functionality. The weights of the model and the tokenizer, \texttt{ gpt-neox-20b} \citep{black2022gpt}, were both imported from Hugging Face \citep{wolf2020huggingfacestransformersstateoftheartnatural}. To obtain the effect of knockout on the Transformer-based GPT-2, we used the implementation published by \citet{geva2023dissecting}.

\begin{figure*}[t]
    \centering
    \includegraphics[width=1\linewidth]{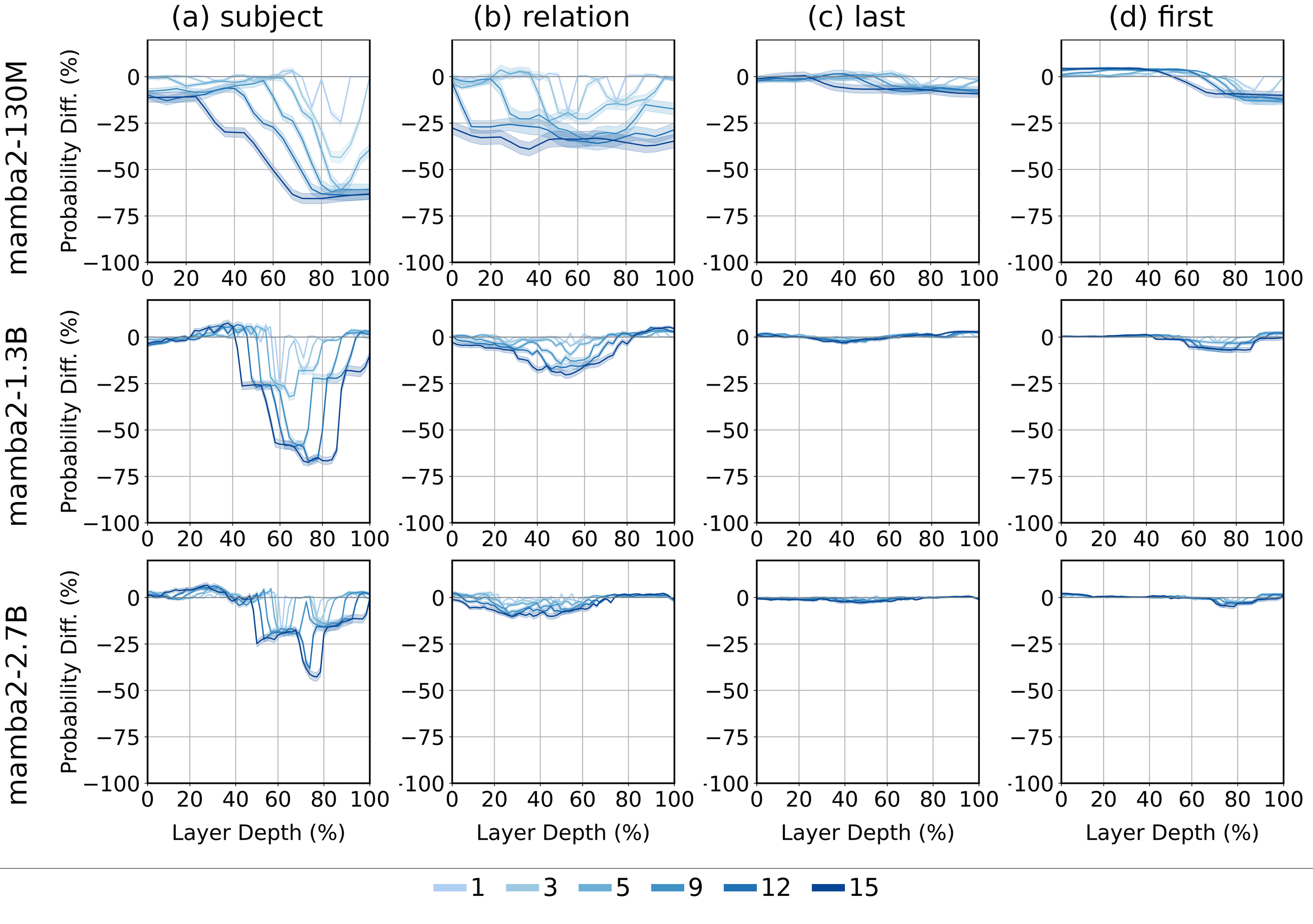}
        \caption{Relative change in correct-token probability after removing the information flow to the final token from various source tokens (columns) on Mamba‑2 with various model sizes (rows) across different window sizes (lines). This figure, which parallels \cref{fig:info_flow_to_last_token}, displays performance for window sizes $\in\{1,3,5,9,12,15\}$. The qualitative patterns observed for Mamba-2 remain consistent, with larger window sizes amplifying these effects. For additional details, see \cref{fig:info_flow_to_last_token,app:implementation}.}
    \label{fig:window_size_ablation_mamba2}
\end{figure*}

\begin{figure*}[t]
    \centering
    \includegraphics[width=1\linewidth]{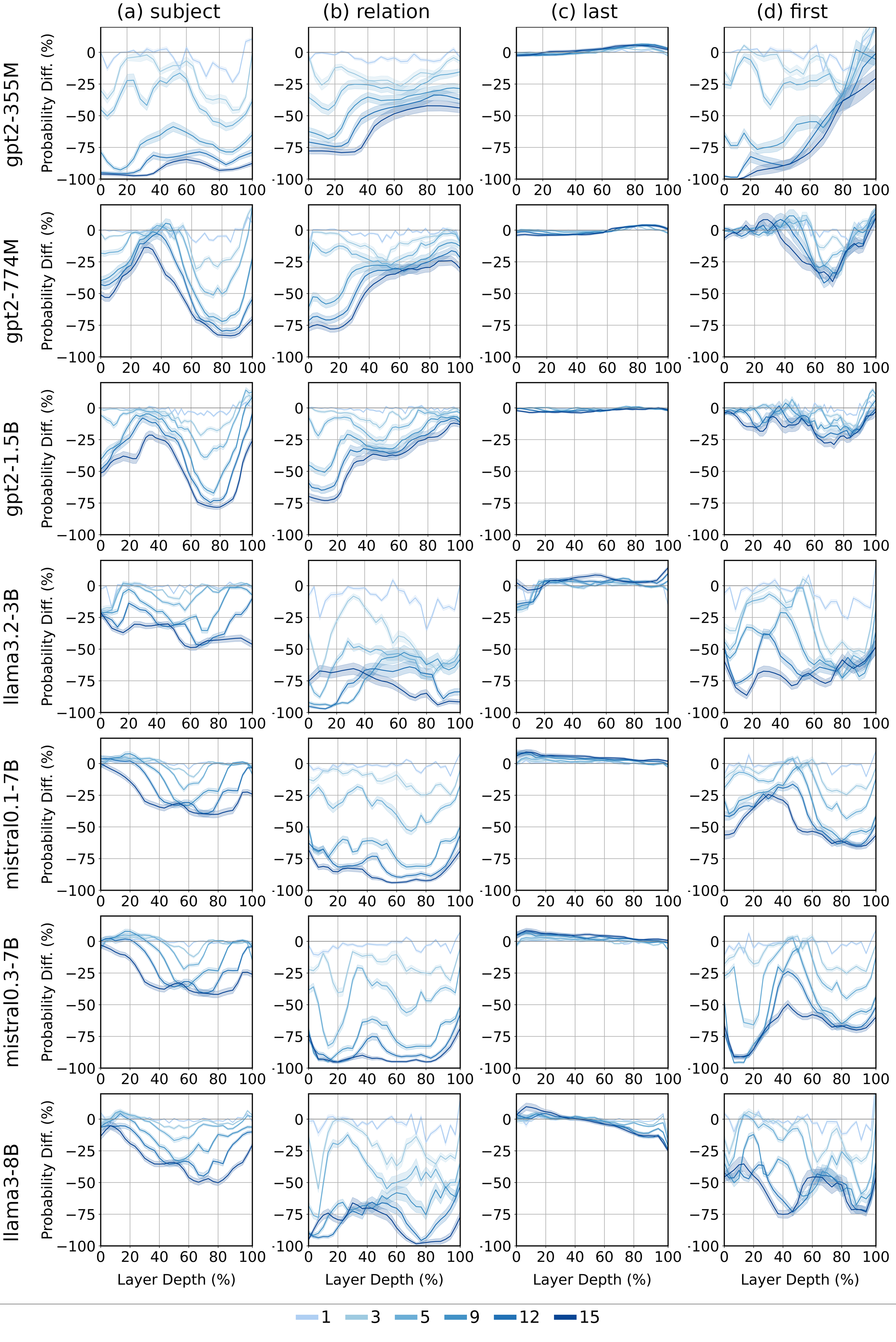}
        \caption{Identical analysis and layout to \cref{fig:window_size_ablation_mamba2}, but applied to Transformer baselines}
    \label{fig:window_size_ablation_transformer}
\end{figure*}

\end{document}